\documentclass{article}

\usepackage[final]{nips_2016}
\usepackage[utf8]{inputenc} 
\usepackage[T1]{fontenc}    
\usepackage{hyperref}       
\usepackage{url}            
\usepackage{booktabs}       
\usepackage{amsfonts}       
\usepackage{nicefrac}       
\usepackage{microtype}      
\usepackage{nips_2016}
\usepackage{times}
\usepackage{latexsym}
\usepackage{dirtytalk}
\usepackage{float}
\usepackage{graphicx}
\usepackage{amssymb}
\usepackage[export]{adjustbox}
\usepackage{xcolor}

\title{Automatic Taxonomy Generation - A Use-Case in the Legal Domain}

\author{Cécile Robin, James O'Neill and Paul Buitelaar \\ \\
	Insight Centre for Data Analytics, \\
	IDA Business Park \\ 
	H91 AEX4 Galway, Ireland \\
   \{james.oneill,cecile.robin,paul.buitelaar\}@insight-centre.org
   }

\date{28-06-2017}

\begin{document}

\maketitle
\section{Abstract}
A key challenge in the legal domain is the adaptation and representation of the legal knowledge expressed through texts, in order for legal practitioners and researchers to access this information easier and faster to help with compliance related issues. One way to approach this goal is in the form of a taxonomy of legal concepts. While this task usually requires a manual construction of terms and their relations by domain experts, this paper describes a methodology to automatically generate a taxonomy of legal noun concepts. We apply and compare two approaches on a corpus consisting of statutory instruments for UK, Wales, Scotland and Northern Ireland laws.

\section{Introduction}
Quicker understanding and comprehension of legal documents is an imperative for practitioners in the legal sector, who have witnessed a steep increase in legislation since the financial crisis in 2008. This can result in law containing even more ambiguous and complex expressions. This can subsequently lead to damaging non-compliance problems for financial institutions. A fundamental way of arranging the knowledge in legal texts to mitigate these problems is by representing the domain in the form of a taxonomy of legal concepts.

Our proposed approach tackles this issue through the automatic construction of a legal taxonomy directly from the content in a legal corpus. The idea here is to be able to create a classification based on the field of application of any type of legal documents, and facilitating the maintenance of the versions. This would help to track changes in regulations and to keep up-to-date with new ones, making this information easily searchable and browsable for Subject Matter Experts (SME).  

We compare two systems for automatic taxonomy generation applied to a small corpus of legal documents. First, we provide related work on automatic taxonomy generation in general, and in the legal domain in particular. We then describe the two approaches chosen for our study. Next, we examine the experiments performed with both systems on a subset corpus of the UK Statutory Instruments, providing a comparative analysis of the results, before providing suggestions for future work.

\section{Related Work}

\paragraph{Generic domain approaches}
Taxonomy cojnstruction is a relatively unexplored area, however \cite{bordea2016semeval} organised a related task in SemEval-2016: TExEval, where the aim was to connect given domain-specific terms in a hyperonym-hyponym manner (relation discovery), and to construct a directed acyclic graph out of it (taxonomy construction). Only one out of the 6 teams produced a taxonomy, focusing thus more on the relation discovery step. Most systems relied on WordNet \cite{fellbaum1998wordnet} and Wikipedia resources.  

\cite{sujatha2011taxonomy} did a structured review of all the main types of approaches involved in the task of automatic taxonomy construction. It includes the use of WordNet, Natural Language Processing (NLP) techniques, tags from web resources, or large external corpora. However, WordNet is a generic lexical resource and is not fitted for the legal language whose definitions and semantic relations are very specific to the domain, as well as constantly evolving. As for external annotated data, these are often non available and also non dynamic resources, therefore not well suited for our task. 

Ahmed et al. \cite{ahmed2012timeline} have used Dynamic Hierarchical Dirichlet Process to track topics over time, documents can be exchanged however the ordering is intact. They also applied this to longitudinal \textit{Neural Information Processing Systems} (NIPS) papers to track emerging and decaying topics (worth noting for tracking changing topics around compliance issues).

Pocostales \cite{pocostales2016nuig} described a semi-supervised method for constructing a \textit{is-a} type relationship (i.e hypernym-hyponym relation) that uses \textit{Global Vectors for Word Representation} (GloVe) vectors trained on a Wikipedia corpus. The approach attempts to represent these relations by computing an average offset for a set of 200 hypernym-hyponym vector pairs (sampled from \textit{WordNet}). This offset distance is then added to each term so that hypernym-hyponyms relations could be identified outside of the 200 pairs which are averaged. In contrast to one of the approaches we present here which also uses word embeddings, we do not restrict the relation to hypernym-hyponym relationships based on WordNet, and therefore this estimated offset for this relation is not required and is not suitable for a taxonomy of legal concepts that are less grounded than that focused on by Pocostales ~\cite{pocostales2016nuig}. 

\paragraph{In the legal domain}
Most work on taxonomy generation in the legal domain has involved manual construction of concept hierarchies by legal experts \cite{buschettu2015kanban}. This task, besides being both tedious and costly in terms of time and qualified human resources, is also not easily adaptable to changes. Systems for automatic legal-domain taxonomy creation have on the contrary received very low attention so far. Only \cite{ahmed2002system} worked on a similar task, and developed a machine learning-based system for scalable document classification. They constructed a hierarchical topic schemes of areas of laws and used proprietary methods of scoring and ranking to classify documents. However, this work has been deposited as a patent and is not freely available. 

We introduce here two methodologies, based on NLP and clustering techniques.

\section{Automatic Taxonomy Construction}

This section describes the two presented bottom-up approaches to taxonomy generation. We begin with a brief description of how noun phrases are extracted followed by an overview of \textit{Hierarchical Embedded Clustering}.
\subsection{Hierarchical Embedded Clustering} 
Hierarchical Embedding Clustering (HEC) is an agglomerative clustering method that we have used for encoding noun phrase predict vectors (i.e Skipgram trained vectors). 

\begin{figure}[h]
\centering
\includegraphics[width=0.8\textwidth]{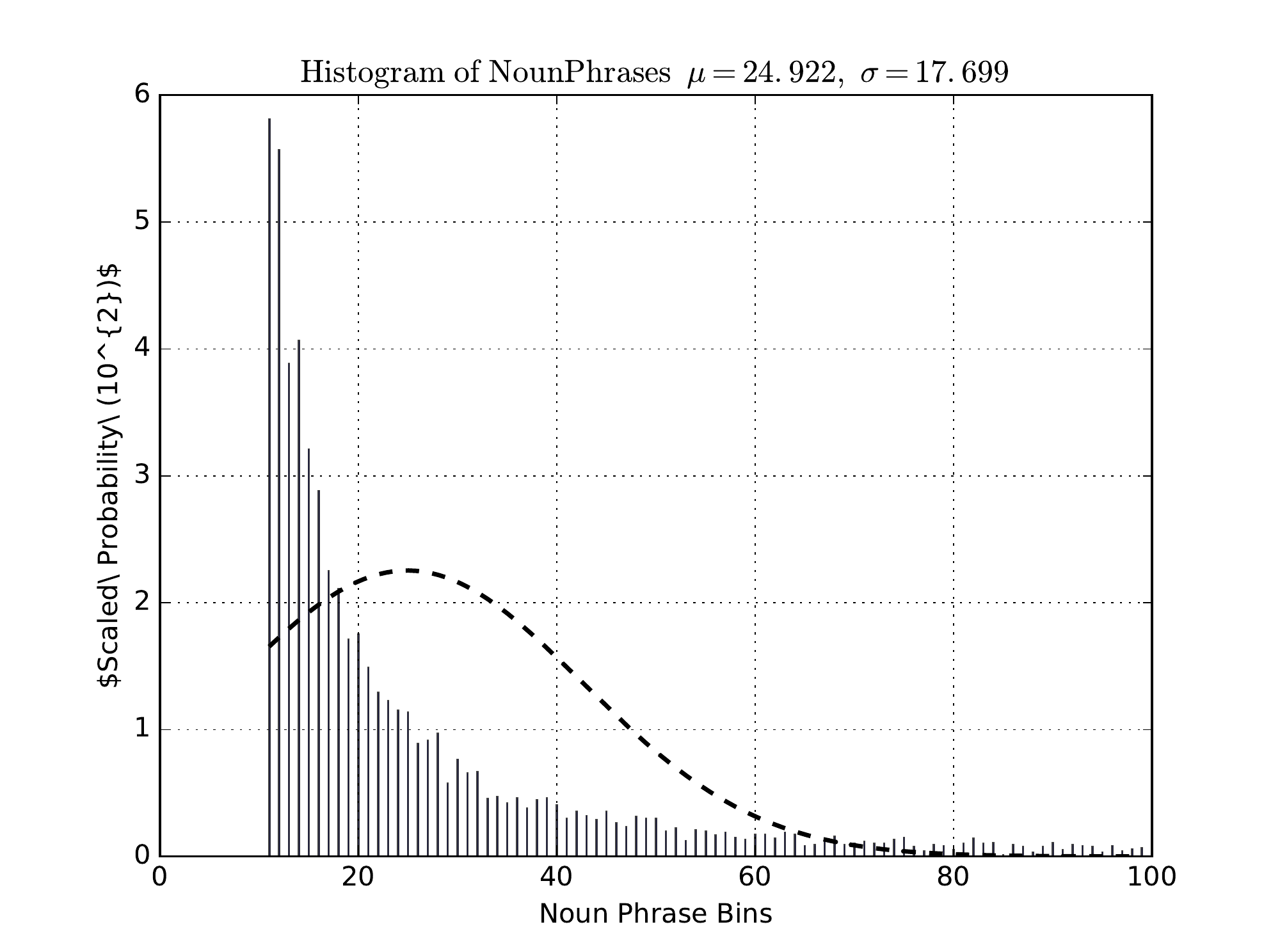}
\caption{Noun Phrase Distribution}
\label{fig:dist_nps}
\vspace{-1em}
\end{figure}

Before performing \textit{HEC} a filtering phase is performed on the corpus to clean potential noisy legal domain syntax (such as the references to regulations e.g. "\textit{Regulation EC No. 1370/2007 means Regulation 1370/2007 ... }" which is not meaningful in our case. 
We then identify noun phrases in the text by extracting bigrams and retaining only the pairs that contain nouns, determined by the NLTK Maximum Entropy PoS tagger\footnote{\url{https://textblob.readthedocs.io/en/dev/\_modules/textblob/classifiers.html}}. This is followed by a filtering stage, whereby the top \textit{n=5377} noun phrases are chosen, based on the highest Pointwise Mutual Information (PMI) scores within a range chosen through a distributional analysis as shown in Figure \ref{fig:dist_nps}. In this figure we present the scaled probability distribution ($10^{2}$) between noun phrase counts in the range $[10-100]$. The dashed line indicates the density, showing that most probability density is lying within the range $[10-60]$. This is a well established trend known as \textit{Luhn's law} ~\cite{pao1978automatic}, where the words with most resolving power lie within the range of low frequencies and high frequencies. Thus, we choose a filtering range between 10-150 to allow for good coverage but still allowing for some domain specificity, resulting in $n=5377$ filtered words and phrases. 

Once noun phrases are selected, we obtain their continuos $\mathtt{skipgram}$ embedded vectors. Each word embedding within a noun phrase is averaged column-wise, therefore representing a whole noun phrase as a single $300$ dimensional vector. This was carried out using both the corpus trained legal vector representations and the large scale pretrained vectors provided by GoogleNews\footnote{\url{https://code.google.com/archive/p/word2vec}} as domain specific legal words are not contained within the GoogleNews vocabulary. 
However, we find that since the legal corpus was relatively small in comparison to Google vectors, it did not achieve the same coherency in grouping noun phrases in a hierarchical structure. Therefore, we focus only on results provided by Google’s pretrained vectors. HEC is a bottom-up approach for creating taxonomies in the sense that each noun phrase is considered as its own cluster at the leaves, which are then incrementally merged until we arrive at the root. There are various linkage methods (e.g single, complete, average etc.) and distance metrics (Euclidean, Manhattan, Hamming, Cosine) for this merging step. We test combinations of the aforementioned linkage and distance measures using Cophenetic Correlation (CC) coefficient $c$ for determining the optimal parameters ($c=1$ means clustering preserves original distances perfectly). CC measures how close the dendrogram is to the pairwise distances between noun phrase embeddings in the corpus. Let instance $x_i$ and $x_j$ be from cluster $C(k_1)$ and $C(k_2)$ in the hierarchy respectively, where $(C(k_1),C(k_2)) \in P(k)$, then the linkage distance $d(x_i,x_j)$ is that of $d(k_1,k_2)$. This is formally in the euclidean form shown in \ref{eq:cc}. Here $X_{(i,j)}$ denotes the distance between $x_i$ and $x_j$, where $\bar{X}$ is the average distance. Similarly, the copehentic distance between any 2 instances $i$ and $j$ is given as $Z_{(i,j)}$.

\begin{equation}\label{eq:cc}
	c = \frac{\sum_{i<j}(X_{(i,j)}-\bar{X})(Z_{(i,j)}-\bar{z})}{\sqrt[]{[\sum_{i<j}(X(i,j)-\bar{X})^{2}][\sum_{i,j}(Z_{(i,j)}-\bar{Z})^{2}]}}
\end{equation}
We find the best results are obtained using Complete Linkage with Euclidean distance, producing $c = 0.612$. Surprisingly using cosine distance from merging embedding clusters performed significantly worse with $c=0.217$ when the euclidean formulation is converted to one of cosine similarity. 

\subsection{Saffron Hierarchy Generation} 
\textit{Saffron} is a software tool\footnote{\url{http://saffron.insight-centre.org/}} that aims to automatically construct a domain-specific hierarchy using domain modeling, term extraction and taxonomy construction. 

\paragraph{Domain Modelling} In order to define the domain of expertise of the corpus , \textit{Saffron} first builds a domain model, i.e. a vector of words representing the highest level of generality in this specific domain \cite{bordea2013phd}. Candidate terms are first extracted using feature selection: giving more weight on part-of-speech carrying meaning, and selecting single words (for genericity) represented in at least a 1/4 of the corpus (for enough specificity to the domain). In order to filter the candidate words, \cite{bordea2013phd} evaluates the coherence of a term within the domain based on \cite{mimno2011optimizing}'s work on topic coherency, following the assumption that domain terms are more general when related to many specific ones. The domain model created is then used in the next phase for the extraction of topics which will make up the taxonomy. 

\paragraph{Topic Extraction} In the topic extraction phase, intermediate level terms of the domain are sought, as defined in \cite{buitelaar2013domain}. It involves two approaches: (1) choosing domain model words in the context of candidate terms (within a defined span size), and (2) using the domain model as a base to measure the lexical coherence of terms by PMI calculation. At the end of this phase, all domain-specific topics are extracted from the corpus, ready to be included in the taxonomy.

\paragraph{Taxonomy Construction} Building connections between the extracted topics is the next step toward the taxonomy construction. Edges are added in the graph for all pairs appearing together in at least three documents, and a generality measure allows to direct edges from generic concepts to more specific ones. A branching algorithm for the construction of domain taxonomies is used \cite{navigli2011graph} to reduce noise in the directed graph of less salient connections. This produces a tree-like structure where the root is the most generic topic, and the topic nodes are going from generalizable parent concepts to more specific downstream concepts.

\subsection{Model Comparison}

The two approaches show similarities and dissimilarities. While both systems use a basic term extraction approach for the selection of candidate noun phrases, and PMI for ranking and filtering them, their approach is different. \textit{Saffron} applies PMI to calculate the semantic similarity of the terms to a domain model, while \textit{HEC} uses the outcome of Luhn's cut analysis instead.
As for taxonomy construction, both methods construct abstract and loosely related connections for the taxonomy hierarchy, instead of the traditional \textit{is-a} relation type. However, \textit{Saffron} defines a global generality measure using PMI to calculate how closely related a term is to other terms from the domain, following the assumption that generic terms are most often used along with a large number of specific terms. On the contrary, \textit{HEC} relies on agglomerative clustering to detect these relations among embedded vector noun phrases, using cosine similarity as similarity measure. For this step, \textit{Saffron} focuses rather on the hierarchy structure at the document level across the texts, whereas \textit{HEC} works directly on all texts within the corpus. This results in abstract concepts at the intermediary levels of the clustering algorithm, and groupings of noun phrases at the leaves. In contrast, \textit{Saffron} provides expressions from these groups at all levels, from the root to the tree.

\section{Experimental Setup}
This section gives a brief overview of the corpus used in our experiments.  
The experiments described here are a first step toward the larger objective of generating a taxonomy for legal corpora over a long time scale. We chose to test the two aforementioned approaches first on a subset of the available Statutory Instruments of Great Britain\footnote{\url{http://www.legislation.gov.uk}}. 41,518 documents have been produced between 2000 and 2016, each year being split in between UK, Scotland, Wales and Northern Ireland. For this experiment, we refine the analysis by selecting the most recent texts (i.e. 2016) of the UK Statutory Instruments (UKSI), that is 838 documents. We don't consider metadata (such as subject matters, directory codes) as are not always available in legal texts. Furthermore, there is no standard schema definition for describing legal documents across different jurisdictions. Our main goal is to compare the results provided by the two different techniques and determine which is the most suitable for the needs described earlier, and focusing on the 2016 UK Statutory Instruments corpus eases the comparison towards that objective.

\section{Results}
\begin{figure}
 \centering
 \includegraphics[width=0.8\textwidth]{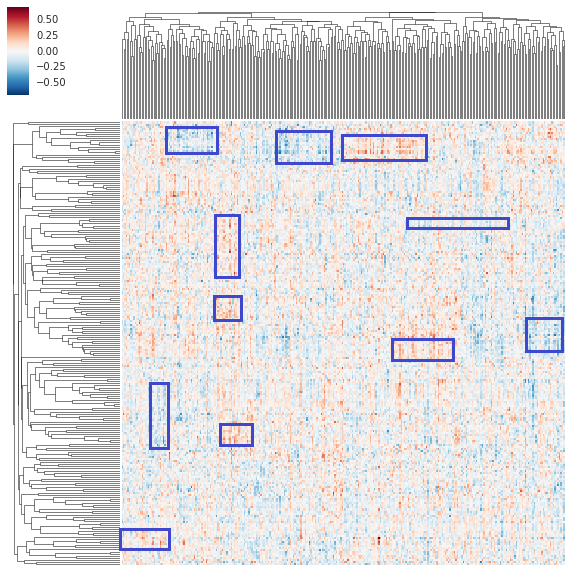}
 \caption{Heatmap of Noun Phrase Vectors (Left Dendrogram = Noun Phrase Hierarchy, Top Dendrogram = Noun Phrase Vector Dimension Hierarchy) \iffalse shows noun phrases on rows  and embedding dimensions on columns)\fi }
 \label{fig:heatmap}
\end{figure}

\paragraph{Hierarchical Clustering Approach}
Figure \ref{fig:heatmap} displays the overall results of \textit{HEC} in the form of a heatmap where noun phrases (rows) and embedding dimension values (columns) are displayed with intensities describing the value of each embedding dimension.  Here the blue bounding box regions represent areas we find in the embedding space that are highly correlated with each other. 
  
The embedded dimensions are reduced representations of the words in an embedding space. Therefore, if the same dimensions of a noun phrase pair both have positively or negatively correlated values in particular dimensions, it means their context is similar in those elements of the vector, meaning that the two noun phrases are related within that given context.
From this figure, it can be identified that some noun phrases are merged due to a small number of dimensions being highly correlated in the embedding space, and not necessarily that all dimensions correlated consistently. This means that the noun phrases are very related only in certain contexts, based on the \textit{GoogleNews} corpus which these vectors have been trained on, but not necessarily appearing together in other contexts. The rectangles within the heatmap aims at pointing out areas within the graph where this is particularly evident.

\begin{figure}
 \centering
 \includegraphics[width=0.5\textwidth]{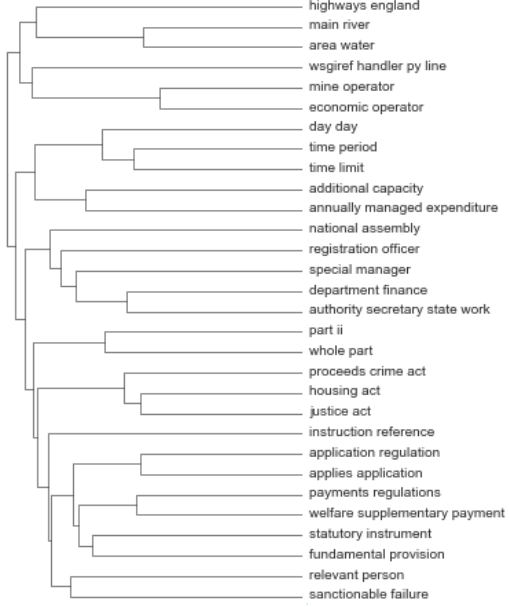}
 \centering
\caption{Sample of \textit{HEC} for UK Statutory Instruments}
 \label{fig:dendro}
\end{figure}

Figure \ref{fig:dendro} shows a snapshot of the results obtained from the previous visualization. Here we can see some interesting groups based on semantic relatedness. The \textit{crime act} and \textit{housing act} have merged with \textit{public interest} and \textit{right of way}, which illustrates a topic within the corpus. Likewise, \textit{mine operator} and \textit{economic operator} have been combined with \textit{merchant shipping}. This appears to show an organized relationship of these two noun concepts.

\begin{figure}
 \centering
 \includegraphics[width=\textwidth]{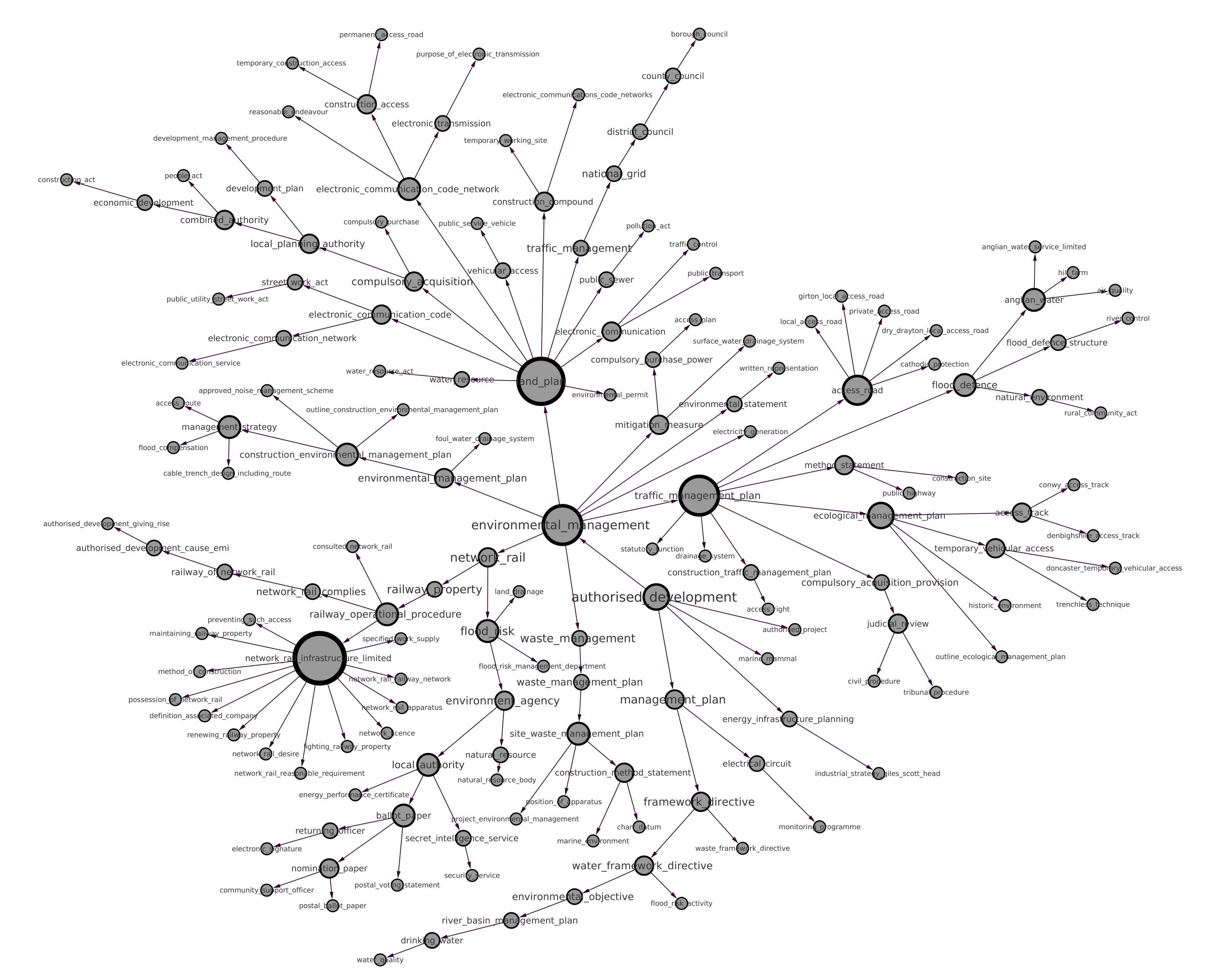}
 \caption{Saffron taxonomy for 2016 UK Statutory Instruments}
 \label{fig:saffron_top}
\end{figure}

\begin{figure}[H]
 \centering
 \includegraphics[width=0.6\textwidth]{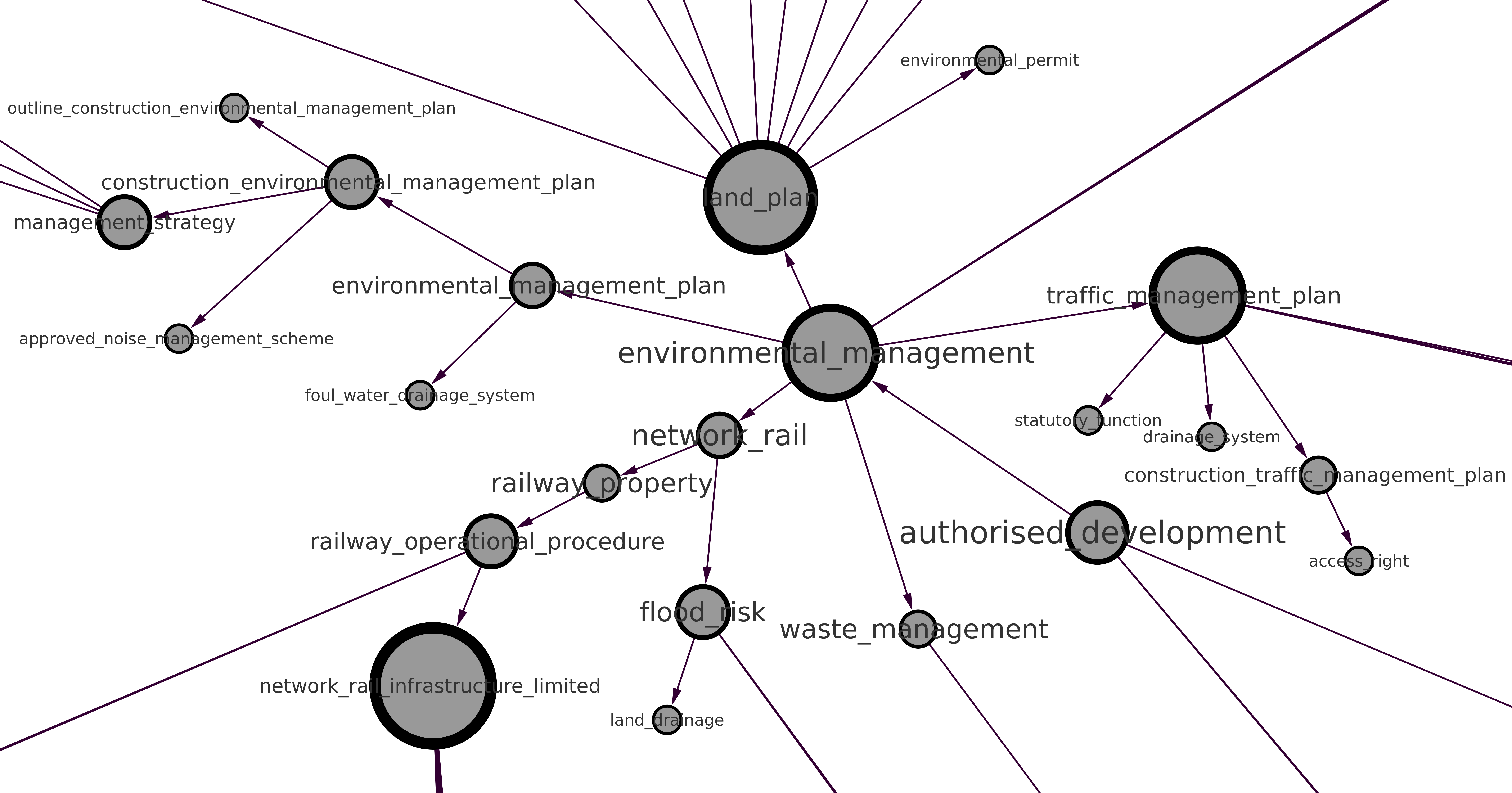}
 \caption{Main topics from 2016 UK Statutory Instruments}
 \label{fig:main_topics}
\end{figure}

\paragraph{Saffron Approach}
We visualize the representation of the taxonomy using an open source software platform, Cytoscape \footnote{\url{http://www.cytoscape.org/}}. Nodes are topics, and the size of the nodes relates to the number of connections each topic shares with others.
Figure \ref{fig:saffron_top} illustrates the whole taxonomy generated by \textit{Saffron} for the corpus. Based on this representation, we detect the topics that are the most prominent in the 2016 UK Statutory Instruments, with four major themes shown in more detail in Figure \ref{fig:main_topics} (\textit{network rail infrastructure limited}, \textit{land plan}, \textit{environmental management} and \textit{traffic management plan}), included in their clusters of related topics. 

Here we can see the directed acyclic graph (DAG) after merged branches from generic concepts to more specific ones hierarchical structure of the graph. For example, the \textit{environmental management} node links to \textit{environmental management plan}, itself redirecting to \textit{construction environmental management plan}, as we can see in Figure \ref{fig:main_topics}.

\begin{figure}[H]
 \centering
 \includegraphics[width=0.6\textwidth]{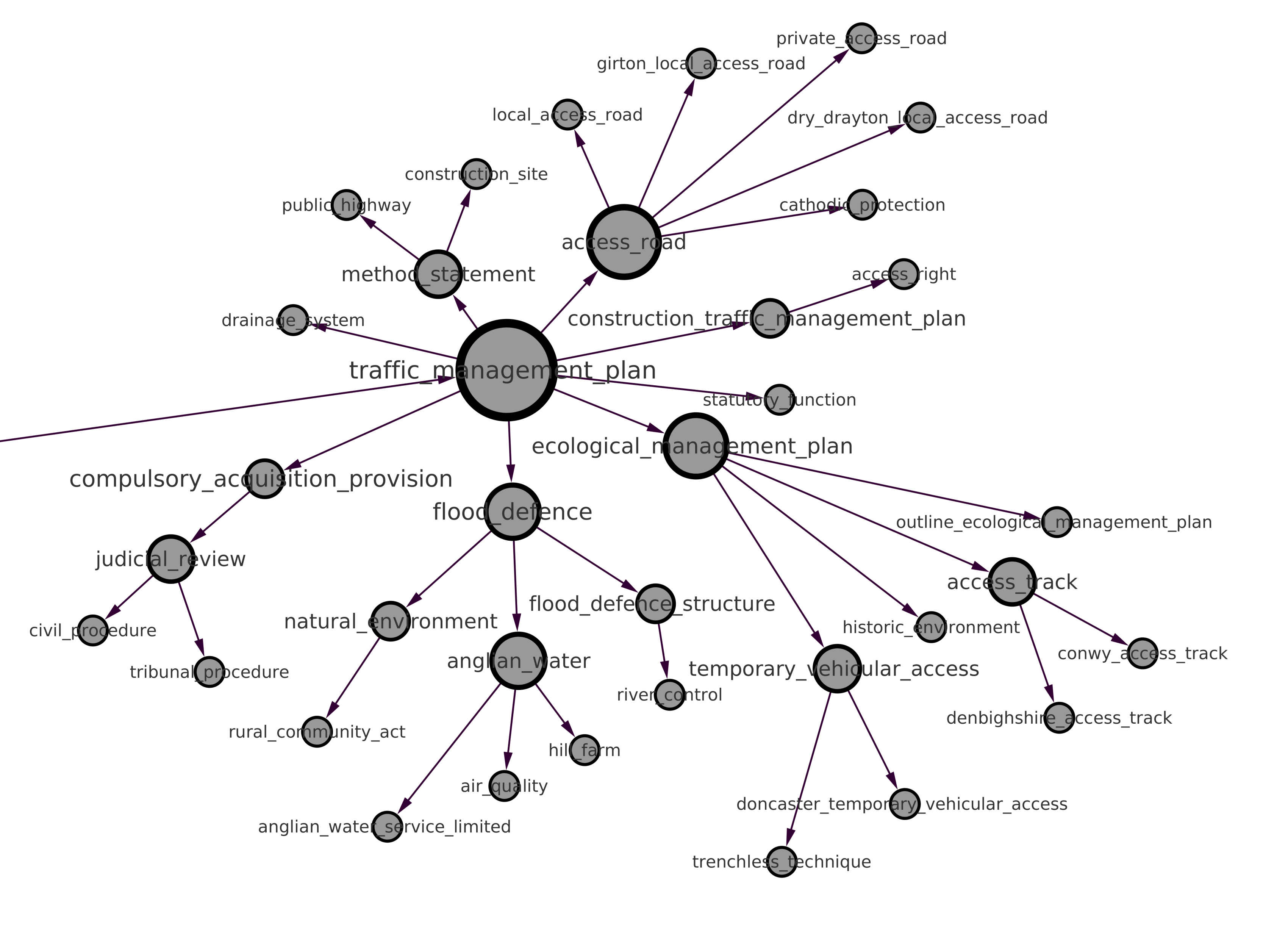}
 \caption{\textit{Traffic Management Plan} topic within the UK Statutory Instruments}
 \label{fig:traffic_management}
 \vspace{-1em}
\end{figure}

There is also a clear interest arising from connecting topics that appear together across the documents. This enables relations between concepts which might not be obvious to a legal practitioner. It would require carrying out an extensive amount of reading within a particular jurisdiction, while still being able to track links between various documents. For example, in Figure \ref{fig:traffic_management}, we observe that \textit{traffic management plan} is connected within some regulation about \textit{drainage system}, the accessibility of the road (\textit{access road)}, and is mentioned through the documents alongside with concepts of \textit{ecological management plan}, and \textit{flood defense structure}. This clearly shows the potential of such semantic processing as an assistance to legal practitioners to identify topics surrounding certain legal issues or for summarizing a whole jurisdiction.

\paragraph{Comparison}
Both of the tree representations display connections between concepts in a different manner. \textit{Saffron} generates a taxonomic structure that is based topical expressions, however HEC focuses on hierarchically clustered word embedding similarities. One can argue that both systems highlight different aspects of the legal domain from the same corpus, and allow to detect different relationships between legal concepts that can be useful for a domain expert. Both methodologies use multi-word expressions to preserve the meaning of expressions that have quite different meaning from the constituent words (e.g \say{Bank of Ireland} different from \say{Bank} and \say{Ireland}).

\section{Conclusion and Future Work}

This work has presented a comparison of two fully automated approaches for identifying and relating salient noun concepts in a taxonomy for the legal domain. The results show coherent groupings of words into legal concepts in both approaches, providing highlights on the emerging topics within the legal corpus. This motivates further research for automatic taxonomy construction to assist legal specialists in various applications. This kind of content management in the legal domain is essential for compliance, tracking change in law and terminology and can also assist legal practitioners in search. 

Although both approaches seem to show interesting results in automatic taxonomy construction, there is a considerable difficulty in evaluating such systems in a quantitative way, due to the lack of benchmarks to evaluate taxonomies created for specific domains, and the low agreement between experts on fast changing areas. In \cite{bordea2016semeval}, the authors evaluated expert agreement on the hierarchical relations between terms. The lowest was shown to be in the Science domain, highlighting the difficulty for experts to get a good overview of a domain which is subject to constant changes. Moreover, their approach to automatically evaluate the resulting hierarchies uses a gold standard taxonomy strictly extracted from \textit{WordNet} \cite{fellbaum1998wordnet}. This resource is too generic for the intermediate level of terms, on which we are focusing in this approach, specifically to the legal domain (eg. \say{notice of appeal}, \say{housing allowance}, \say{pension scheme}). 
However, we plan on carrying out further studies towards a formal representation of concepts within a domain, undertaken by domain experts. This kind of benchmark would establish an evaluation dataset for this domain, where the generated taxonomies are evaluated with taxonomy matching and alignment measures. We also consider establishing an expert user study to evaluate the generated results, with the idea to get legal domain practitioners' views on the practicability of such representation, and the pertinence of the relations established.

\bibliography{nips_2016}
\bibliographystyle{nips_2016}

\end{document}